\documentclass[conference]{IEEEtran}
\IEEEoverridecommandlockouts

\usepackage[noadjust]{cite}

\usepackage{amsmath,amssymb,amsfonts}
\usepackage{algorithmic}
\usepackage{graphicx}
\usepackage{textcomp}
\usepackage{xcolor}
\usepackage[pagebackref=true,breaklinks=true,letterpaper=true,colorlinks,bookmarks=false]{hyperref}
\usepackage{amsmath}
\usepackage{amssymb}
\usepackage{graphics}
\usepackage{booktabs}
\usepackage{multirow}
\usepackage{rotating}
\usepackage{url}
\usepackage{mathrsfs}

\usepackage{makecell}
\usepackage{rotating}
\usepackage{color}
\usepackage{caption}
\usepackage{multirow}
\usepackage{booktabs}
\usepackage{hhline}

\usepackage[capitalize]{cleveref}
\crefname{section}{Sec.}{Secs.}
\Crefname{section}{Section}{Sections}
\Crefname{table}{Table}{Tables}
\crefname{table}{Tab.}{Tabs.}

\newcommand{\ie}{\textit{i}.\textit{e}.}
\newcommand{\eg}{\textit{e}.\textit{g}.}
\DeclareRobustCommand{\IEEEauthorrefmark}[1]{\smash{\textsuperscript{\footnotesize #1}}}

\usepackage[utf8]{inputenc}
\usepackage{pifont}
\usepackage{newunicodechar}
\newunicodechar{✓}{\ding{51}}
\newunicodechar{✗}{\ding{55}}

\makeatletter
\newcommand{\linebreakand}{%
  \end{@IEEEauthorhalign}
  \hfill\mbox{}
  \par
  \mbox{}\hfill\begin{@IEEEauthorhalign}
}
\makeatother

\def\BibTeX{{\rm B\kern-.05em{\sc i\kern-.025em b}\kern-.08em
    T\kern-.1667em\lower.7ex\hbox{E}\kern-.125emX}}
    
\begin{document}

\title{CattleEyeView: A Multi-task Top-down View Cattle Dataset for Smarter Precision Livestock Farming\\
}

\author{\IEEEauthorblockN{Kian Eng Ong\IEEEauthorrefmark{1}\textsuperscript{‡}}
\and
\IEEEauthorblockN{Sivaji Retta\IEEEauthorrefmark{2}\textsuperscript{‡}}
\and
\IEEEauthorblockN{Ramarajulu Srinivasan\IEEEauthorrefmark{2}\textsuperscript{‡@}} 
\and
\IEEEauthorblockN{Shawn Tan\IEEEauthorrefmark{2}}
\and
\IEEEauthorblockN{Jun Liu\IEEEauthorrefmark{1}}

\linebreakand
\IEEEauthorblockA{
\normalsize {\IEEEauthorrefmark{1}} Information Systems Technology and Design, 
Singapore University of Technology and Design, Singapore
\\
\normalsize {\IEEEauthorrefmark{2}} AnimalEYEQ Private Limited, Singapore
\\
}

}

\maketitle

\def\thefootnote{{\textsuperscript{‡}}}\footnotetext{Equal contributions. \textsuperscript{@}Corresponding author: {rama@animaleyeq.com}}

\begin{abstract}
Cattle farming is one of the important and profitable agricultural industries. Employing intelligent automated precision livestock farming systems that can count animals, track the animals and their poses will raise productivity and significantly reduce the heavy burden on its already limited labor pool. To achieve such intelligent systems, a large cattle video dataset is essential in developing and training such models. However, many current animal datasets are tailored to few tasks or other types of animals, which result in poorer model performance when applied to cattle. Moreover, they do not provide top-down views of cattle. To address such limitations, we introduce CattleEyeView dataset, the first top-down view multi-task cattle video dataset for a variety of inter-related tasks (\ie, counting, detection, pose estimation, tracking, instance segmentation) that are useful to count the number of cows and assess their growth and well-being. The dataset contains 753 distinct top-down cow instances in 30,703 frames (14 video sequences). We perform benchmark experiments to evaluate the model's performance for each task. The dataset and codes can be found at \href{https://github.com/AnimalEyeQ/CattleEyeView}{https://github.com/AnimalEyeQ/CattleEyeView}
\end{abstract}

\begin{IEEEkeywords}
dataset, cow, detection, tracking, counting, pose estimation, instance segmentation
\end{IEEEkeywords}

\section{Introduction}
Monitoring and assessing the health, well-being, growth, and behaviors of animals are especially important in the livestock farming industry, as all of these ultimately impact farm production and profits \cite{Multi-Scene_Cow_Pose}. Cattle farming is one of the important and profitable agricultural industries, especially in the United States \cite{USDA}. Cows are raised for their meat as well as milk which is then further processed to dairy products (\eg, cheese, butter, yogurt). Globally, the demand for cattle is expected to increase by 5.8\% in the next 10 years \cite{FAO_Projection}. Given the strain on the farming industry with an increasing number of animals but decreasing number of farmers \cite{Livestock_Farming_Intro}, employing automated Artificial Intelligence (AI) systems in the livestock farming industry will raise productivity and expedite the industry's effort in digital transformation towards smart precision livestock farming while significantly reducing the heavy burden on its already limited labor pool \cite{Pig_Pose, Multi-Cow_Pose}. Precision livestock farming includes the use of intelligent systems to count animals, track and analyze poses and movements of animals and their heads, bodies, and limbs, as well as their activities and behaviors in the farm \cite{Animal_Pose_Tracking, Livestock_Farming_Intro}. 

Accurate counting of livestock is particularly important in the livestock industry as each farm on average sells up to 5000 cows each day \cite{NVIDIA}. Counting errors are common when manually counted by humans and are substantially costly, given that each cow is priced at around USD\$1000 \cite{NVIDIA}. 

Besides counting, tracking each animal in the farm is precursor to evaluating its state of health and analyzing its behaviors and interactions with others \cite{WATB, AnimalTrack}. Additionally, detecting its keypoints is important in many practical application scenarios such as animal identification, tracking, behavioral analysis (including lameness detection), part segmentation, and even gait analysis \cite{Quadruped_Animals, Animal_Pose_Survey}. On top of keypoints identification, segmentation can help to chart and assess the animal's well-being (\eg, cleanliness and weight) \cite{Segmentation}. All of these interrelated tasks are crucial to achieve a comprehensive understanding and analysis of the animal's health and well-being. To develop intelligent systems for smart precision livestock farming, a large video dataset is essential in developing and training such models. However, there is a lack of sufficient detailed annotations of animal videos that can be used for training \cite{Animal_Pose_Survey}, which currently impedes research in this field. 

Although there are existing animal datasets, some contain a diversity of animals (\eg, Animal Pose \cite{Animal_Pose}, AP10K \cite{AP10K}, Animal Kingdom \cite{Animal_Kingdom}) while others are for specific types of animals (\eg, sheep \cite{Sheep}, and pigs \cite{Pig_Pose}). Models that are developed on such datasets are unable to achieve good performance when applied to cattle. Moreover, most existing animal datasets provide frontal and side views of the animals, but not the top-down views of animals. The advantages of using the top-down view, instead of frontal or side views, of the animal are \textbf{(1)} the animals are usually not severely occluded by other farm objects or animals, \textbf{(2)} the appearances of animals are more consistent (similar size and shape) as compared to isometric angled view that is commonly seen in CCTV footages \cite{Pig_Pose}, \textbf{(3)} the top-down view focuses on a particular animal, 
unlike isometric view which may contain differently sized animals in the background that will distract the model. This is important to achieve accurate counting of animals as well as taking into account the direction of movement of the animal for accurate counting. All of these advantages are important to accurately assess the growth and well-being of the animal. However, models that are developed on existing datasets based on frontal, side or isometric views of animals are unable to achieve good performance when applied to situations like ours that require the top-down views of cattle. There is only one cattle dataset \cite{Cows2021} with a top-down view that only localizes and re-identifies 186 cattle based on their black-and-white coat patterns viewed from top-down. 

Hence, to address this limitation, we construct CattleEyeView, the \textit{first} multi-task video cattle dataset using the bird's-eye views (\ie, top-down views) of 6 different cattle breeds with different coat colors. Our CattleEyeView dataset not only provides carefully labeled annotations of \textbf{(1)} bounding boxes of the cow's body and head for cattle counting and tracking, we also provide detailed annotations for \textbf{(2)} tracking ID for tracking, \textbf{(3)} cattle count (\ie, number of cattle that have passed through the ramp) in the video sequence for cattle counting, as well as \textbf{(4)} 24 keypoints for pose estimation, and \textbf{(5)} segmentation masks for instance segmentation. Our CattleEyeView dataset with various types of detailed annotations shall facilitate the community and industry to develop, adapt, and evaluate various types of advanced models, thereby promoting the development and deployment of smarter precision livestock farming AI systems to fulfill diverse industry needs and application scenarios. 

\section{Related Works}
Although there are existing animal datasets, some contain a diversity of animals (\eg, Animal Pose \cite{Animal_Pose}, AP10K \cite{AP10K}, Animal Kingdom \cite{Animal_Kingdom}) from which the models are developed on but often do not achieve good performance when applied to specific types of farm animals. On the other hand, most animal datasets are catered for specific task (\eg, pose estimation only \cite{AP10K}), specific types of animals (\eg, sheep \cite{Sheep}, and pigs \cite{Pig_Pose}), or specific environments (\eg, mice in laboratories \cite{Mice}). Unlike controlled environments in a laboratory, farms often have non-uniform background (\eg, dirt ground) and sometimes with varying lighting conditions (especially outdoors). Furthermore, the low contrast between the animal (\eg, cow) and the dirt ground, as well as occlusion of the animals by other animals or even equipment used in the farm, also pose practical challenges.

Moreover, most existing animal datasets provide frontal and side views of the animals, but not the top-down views of animals. More crucially, models that are trained on existing animal datasets with frontal, side or isometric views of animals do not generalize well on top-down view of animals, because of the difference in viewpoint. Therefore, there is a need for such top-down view datasets in order to develop advanced intelligent automated system to cater for different needs in precision livestock farming. However, there is only one cattle dataset with a top-down view. Cows2021 \cite{Cows2021} only localizes and re-identifies 186 cattle based on their black-and-white coat patterns viewed from top-down. As it primarily localizes and re-identifies the cattle, it does not contain annotations for head bounding box, pose, and counting. 

Different from existing animal datasets as shown in \Cref{tab:comparison}, our CattleEyeView video dataset not only provides carefully labeled annotations of bounding boxes for cattle tracking and counting, our annotations also include keypoints for pose estimation, and segmentation masks for instance segmentation. Therefore, our dataset containing six different cow breeds with different coat colors in different illumination conditions provides a challenging benchmark for the research community and industry to develop, adapt, and evaluate various types of advanced methods for smarter precision livestock farming.

\begin{table*}

\caption{Comparison with other cattle datasets}
\vspace{-0.2cm}
\label{tab:comparison}
\resizebox{\linewidth}{!}{
\begin{tabular}{|c|c|c|c|c|c|c|c|c|c|} 
\hline
\multirow{2}{*}{\textbf{Dataset}} & \multirow{2}{*}{\begin{tabular}[c]{@{}c@{}}\textbf{Publicly}\\\textbf{available?}\end{tabular}}

& \multicolumn{5}{c|}{\textbf{Task}} & \multirow{2}{*}{\textbf{\textbf{\# frames}}} & \multirow{2}{*}{\textbf{\textbf{\# keypoints}}} & \multirow{2}{*}{\textbf{Type of view}} \\ 
\cline{3-7}
 &  & \begin{sideways}\textbf{Detection}\end{sideways} & \begin{sideways}\textbf{Counting}\end{sideways} & \begin{sideways}\textbf{Tracking}\end{sideways} & \begin{sideways}\textbf{Pose}\end{sideways} & \begin{sideways}\textbf{Segmentation}\end{sideways} &  &  & \\ 
\hline
Cows2021 \cite{Cows2021} & \checkmark & \checkmark & ✗ & ✗ & ✗ & ✗ & 10,402 & N.A. & Top \\
Deep cascaded convolutional models for cattle pose estimation \cite{DeepCascaded} & ✗ & ✗ & ✗ & ✗ & \checkmark & ✗ & 2134 & 16 & Side \\
Video-based multi-scene cow pose estimation \cite{Multi-Scene_Cow_Pose} & ✗ & ✗ & ✗ & \checkmark & \checkmark  & ✗ & 3000 & 16 & Side \\
Multicow pose estimation based on keypoint extraction \cite{Multi-Cow_Pose} & ✗ & \checkmark  & ✗ & ✗ & \checkmark & ✗ & 1800 & 16 &  Side \\
CoWalk-30 \cite{T-LEAP} & ✗ & ✗ & ✗ & ✗ & \checkmark  & ✗ & 4236 & 17 &  Side \\ 
\hline
\textbf{CattleEyeView (Ours)} & \textbf{\checkmark} & \textbf{\checkmark} & \textbf{\checkmark} & \textbf{\checkmark} & \textbf{\checkmark} & \textbf{\checkmark} & 30,703 & 24 & Top \\
\hline
\end{tabular}
}
\vspace{-0.5cm}
\end{table*}

\section{CattleEyeView Dataset}
\subsection{Video Data Collection and Annotation}
\begin{figure}[ht]
\centering
\includegraphics[width=0.5\columnwidth]{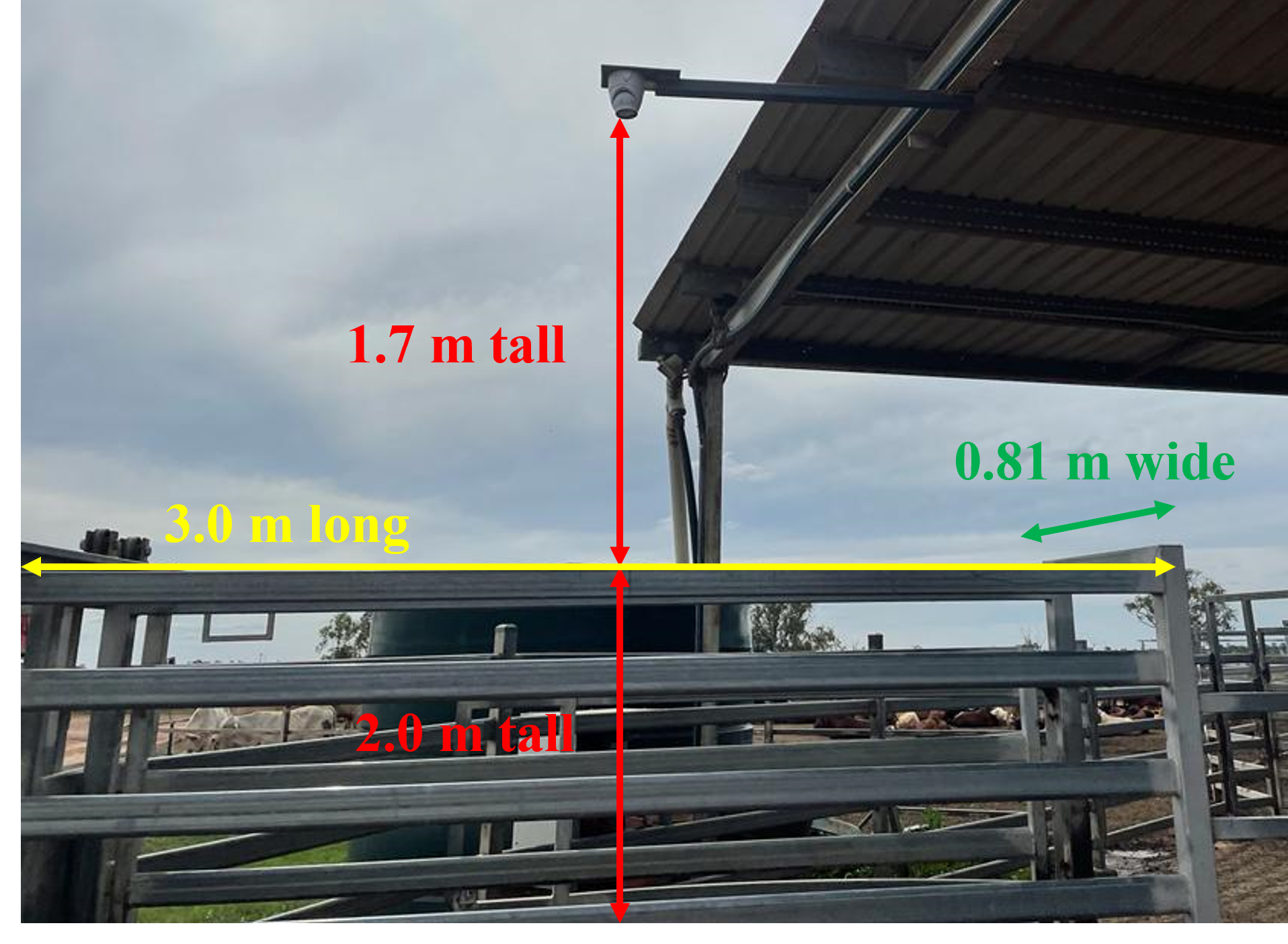}
\caption{The camera is mounted at a height of 3.7m to capture the top-down views of cattle.}
\label{camera}
\end{figure}

\begin{figure}[ht]
\centering
\includegraphics[width=0.8\columnwidth]{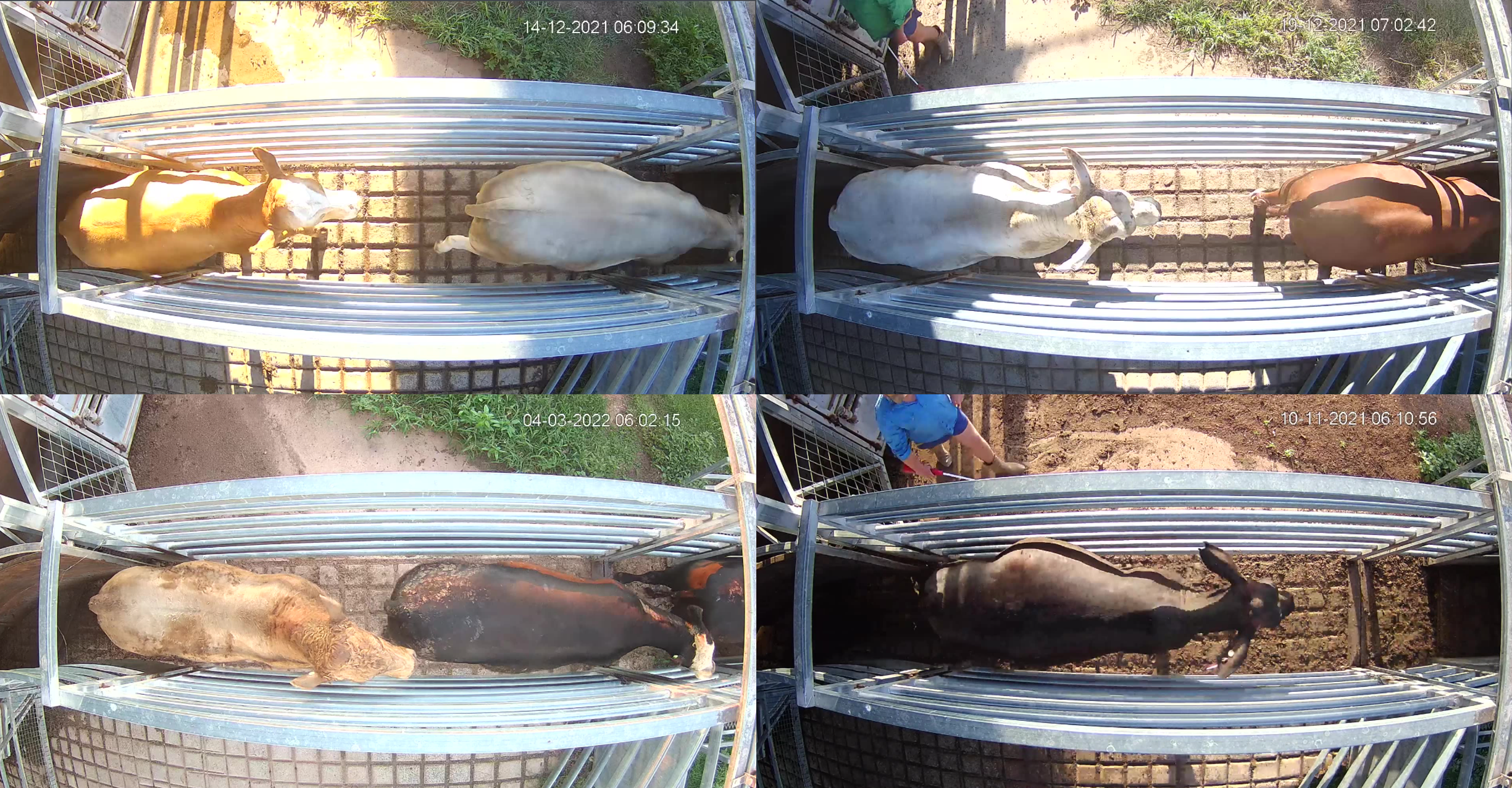}
\caption{The video sequences contain different colored cows found in different lighting conditions (different times and different days).}
\label{scenes}
\vspace{-0.5cm}
\end{figure}

The 14 video sequences (containing 30,703 frames) containing top-down views of cows are collected in the morning and evening over 4 months (2021-11-09 to 2022-03-04) from overhead RGB CCTV cameras fixed at a height of 3.7m with cows walking through the outdoor loading ramp of a farm (\cref{camera}). There cows have different coat colors and patterns, and are seen in different illumination conditions (\cref{scenes}). In total, there are 753 distinct cow instances which are manually annotated by experienced human annotators and subsequently verified by more senior annotators. 

\vspace{-0.1cm}

\begin{figure}[ht]
\centering
\includegraphics[width=1\columnwidth]{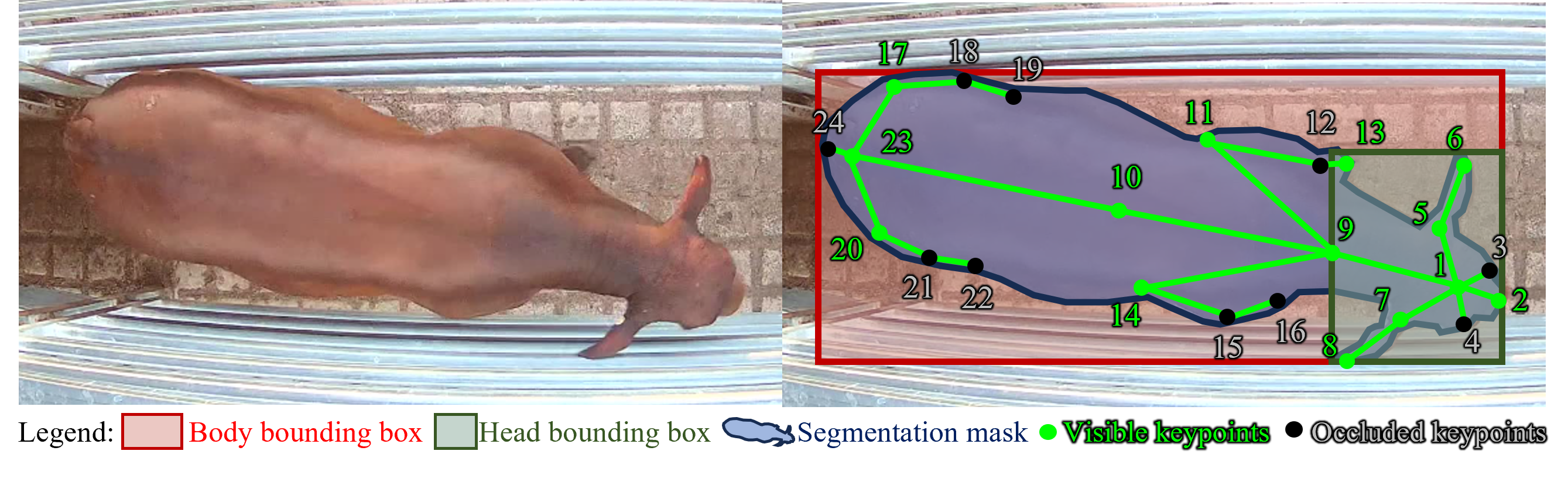}
\caption{Example of annotations in a frame of the video sequence: Bounding boxes of the head and body of the cattle, as well as keypoints and segmentation mask of the cattle. Other annotations not shown in the image above include tracking ID and count number of cattle that have passed through the ramp.}
\label{eg_annot}
\vspace{-0.5cm}
\end{figure}

An example of the annotation can be seen in \cref{eg_annot}. We annotate the body and head bounding boxes, as well as the segmentation mask of the cows. We also track the cows by assigning a unique tracking ID and count the number of cows as they walk through the loading ramp of a farm. 

Additionally, we annotate 24 keypoints, together with the respective visibility tag (\ie, visible and occluded), for the 
\textbf{(1)} head, 
\textbf{(2)} nose, 
\textbf{(3)} left eye, 
\textbf{(4)} right eye, 
\textbf{(5)} base of left ear, 
\textbf{(6)} tip of left ear, 
\textbf{(7)} base of right ear, 
\textbf{(8)} tip of right ear, 
\textbf{(9)} neck, 
\textbf{(10)} withers, 
\textbf{(11)} left of front elbow, 
\textbf{(12)} left of front knee, 
\textbf{(13)} left of front paw, 
\textbf{(14)} right of front elbow, 
\textbf{(15)} right of front knee, 
\textbf{(16)} right of front paw, 
\textbf{(17)} left of back elbow, 
\textbf{(18)} left of back knee, 
\textbf{(19)} left of back paw, 
\textbf{(20)} right of back elbow, 
\textbf{(21)} right of back knee, 
\textbf{(22)} right of back paw, 
\textbf{(23)} base of tail, and 
\textbf{(24)} end of tail. The head keypoints (1 to 10) and rear keypoints (23 and 24) are chosen to reflect the where the cow is facing, and also take into account when some keypoints may be occluded. The limb keypoints (11 to 22) are landmarks to indicate how the cow moves, as well as facilitate the assessment of its growth (including weight and gait).

\subsection{Dataset Description}
The CattleEyeView dataset contains 753 distinct top-down cow instances in 30,703 frames (14 video sequences) with (1) bounding boxes of the cow’s body and head for cattle counting and tracking, we also provide detailed annotations for (2) tracking ID for tracking, (3) cattle count (\ie, number of cattle that have passed through the ramp) in the video sequence for cattle counting, as well as (4) 24 keypoints for pose estimation, and (5) segmentation masks for instance segmentation

\textbf{Top-down cattle view.} The top-down views of animals are usually not severely occluded by farm objects or other animals. Additionally, the appearances of animals in top-down view are more consistent (similar size and shape) than isometric view that is commonly seen in CCTV footages. Moreover, the top-down view focuses on a particular animal, unlike isometric view which may contain differently sized animals in the background that distracts the model. Hence, these advantages are beneficial in accurately counting the animals and analyzing their growth and well-being.

\textbf{Different cow breeds and different coat colors.} Our dataset contains images of six different cow breeds (Angus, Brahman, Charolais, Droughtmaster, Hereford, and Santa Gertrudis). These cows have different coat colors. Having a diversity of cow breeds and coat colors improves the generalization ability of the models to detect the cows and perform various downstream tasks. 

\textbf{Various illumination conditions.} The footages are obtained at different times of days, with different illumination conditions. The shadow cast on the cows sometimes results in low foreground-background contrast, thereby introducing significant and practical challenges such as identifying and delineating the cows and the dirt ground.

\subsection{Dataset Tasks and Annotations}
\begin{figure}[htbp]
\centering
\includegraphics[width=1\columnwidth]{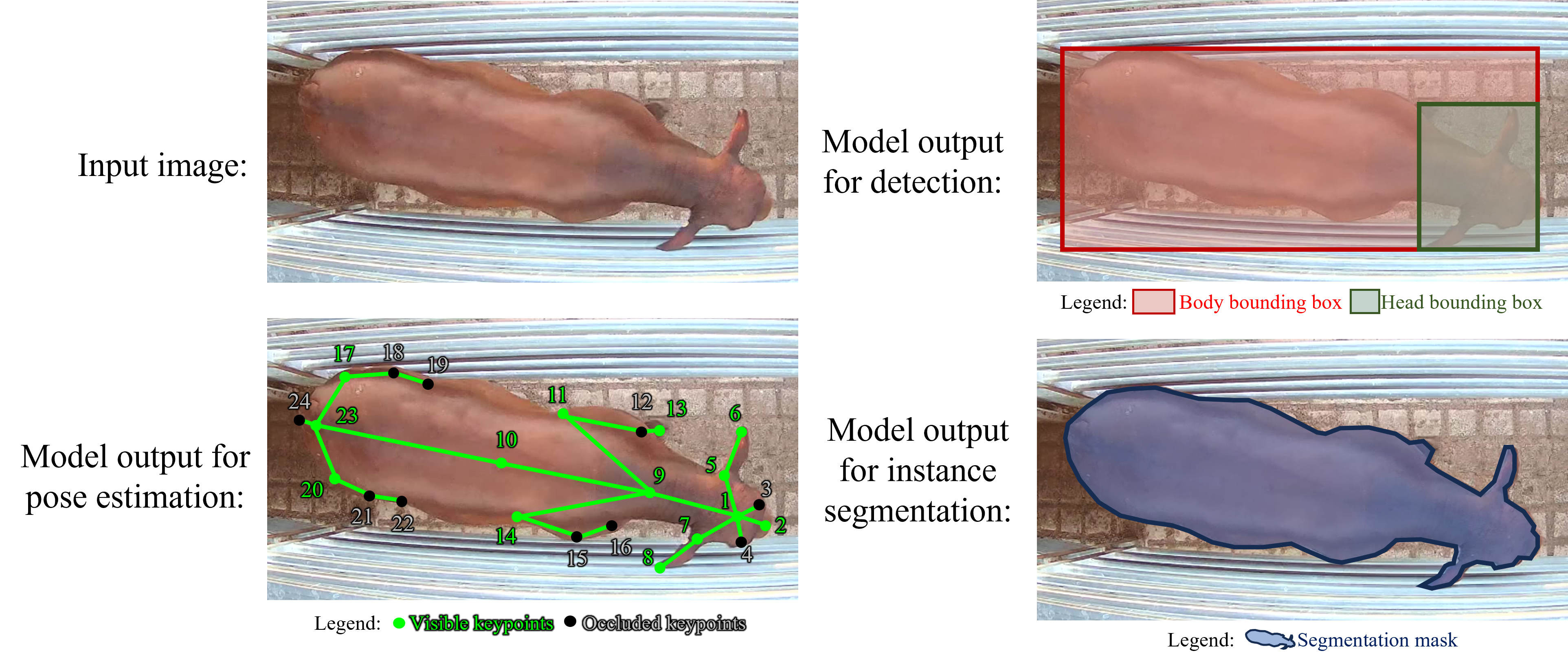}
\caption{In object detection (top-right image), the model identifies and outputs the spatial location of each animal. In pose estimation (bottom-left image), the model predicts all the keypoints of each animal. In instance segmentation (bottom-right image), the model outputs the binary segmentation mask of each animal.}
\label{tasks}
\vspace{-0.3cm}
\end{figure}

\textbf{Object detection.}
In detection, the model takes in the image of the animal and outputs the spatial location (bounding box) of the animal in the frame (\cref{tasks}).

\begin{figure}[htbp]
\centering
\includegraphics[width=1\columnwidth]{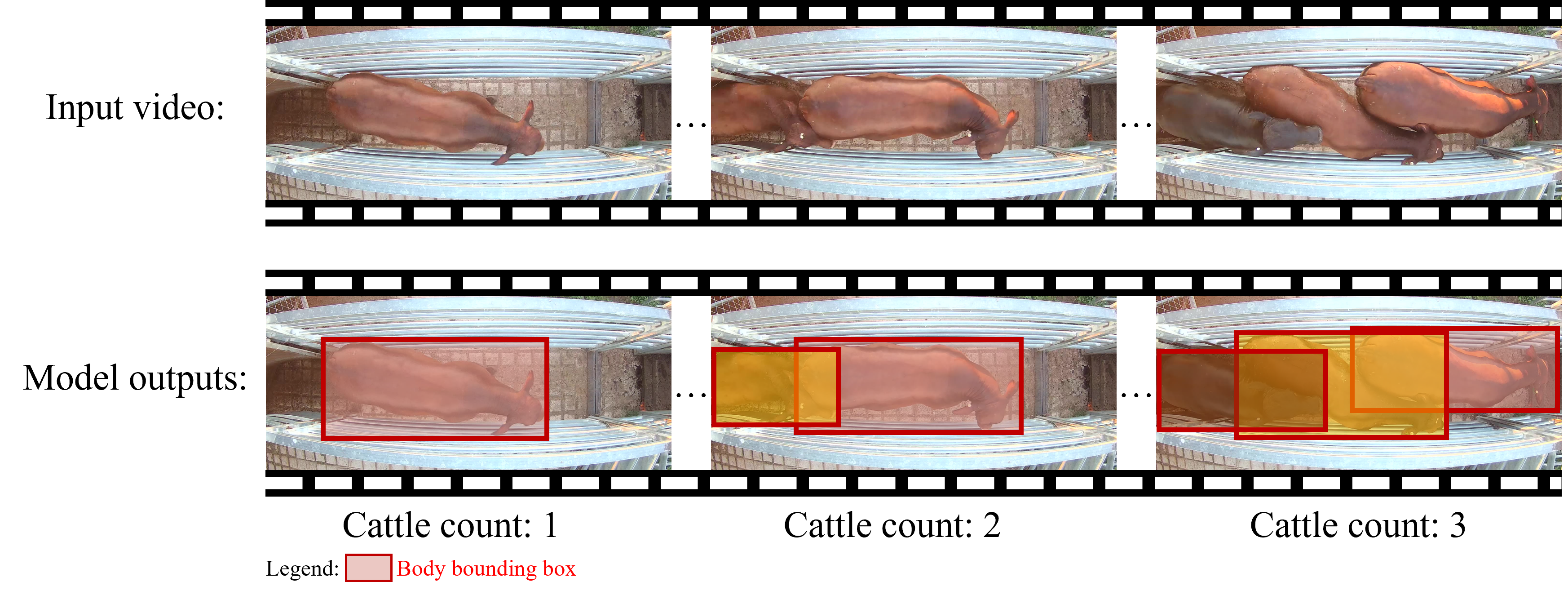}
\caption{In tracking, the model identifies and tracks the movement of each animal.}
\label{tracking}
\vspace{-0.3cm}
\end{figure}

\textbf{Tracking and counting.}
In tracking, the model takes in the video of the animal and predicts the trajectory of the animal (\ie, bounding box across frames) (\cref{tracking}). For counting, the model takes in the video of the animal and outputs the number of animals that have crossed a pre-defined region in the frame of the video (\cref{tracking}).

\textbf{Pose estimation.}
In pose estimation, the model takes in the image of the animal and predicts the different keypoints (\eg, joints) on the animal (\cref{tasks}). 

\textbf{Instance segmentation.}
In instance segmentation, the model takes in the image of the animal and outputs the binary segmentation mask of each animal, which outlines each instance of the animal (\cref{tasks}).

In a nutshell, our CattleEyeView dataset contains various types of annotations that will be useful to train various types of models for smarter precision livestock farming. Our dataset provides a challenging benchmark for the research community and industry to develop, adapt, and evaluate various types of advanced methods for smarter precision livestock farming.

\section{Experiments}
To facilitate training for various tasks, each video is either assigned to the train set (75\%) or test set (25\%). We use YOLOv5 \cite{YOLOv5}, YOLOv8 \cite{YOLOv8}, OpenMMLab \cite{OpenMMLab}, and BoxMOT \cite{BoxMOT} code libraries to evaluate state-of-the-art methods based on respective existing established evaluation metrics on our dataset (YOLOv8 \cite{YOLOv8} for cattle detection, BoxMOT using YOLOv8 with ByteTrack \cite{ByteTrack} and YOLOv8 with BoT-SORT \cite{BoT-SORT} for tracking and counting, HRNet \cite{HRNet} and YOLOv8-Pose \cite{YOLOv8} for pose estimation, YOLOv5-Segment \cite{YOLOv5} and YOLOv8-Segment \cite{YOLOv8} for instance segmentation). 

\Cref{tab:detection} shows the results of the body and head detection. The models achieve high mAP$_{0.5}$ scores for both body and head, despite color variations of the cows and different lighting conditions. The head detection works better than body detection, thus suggesting that further works are needed to improve the body detection. In comparison, the models achieve better performance for Cows2021 dataset, hence suggesting that advanced methods are needed to improve performance on our dataset that is more challenging.

\vspace{-0.2cm}
\begin{table}[ht]
\caption{Results of cattle detection}
\vspace{-0.2cm}
\label{tab:detection}
\resizebox{\linewidth}{!}{
\begin{tabular}{|c|c|cccc|cccc|} 
\hline
\multirow{2}{*}{\textbf{Dataset}} & \multirow{2}{*}{\textbf{Model}} & \multicolumn{4}{c|}{\textbf{Body}} & \multicolumn{4}{c|}{\textbf{Head}} \\ 
\cline{3-10}
 & & \begin{sideways}\textbf{Precision↑}\end{sideways} & \begin{sideways}\textbf{Recall↑}\end{sideways} & \begin{sideways}\textbf{mAP$_{0.5}$↑}\end{sideways} & \begin{sideways}\textbf{mAP$_{0.5:0.95}$↑}\end{sideways} & \begin{sideways}\textbf{Precision↑}\end{sideways} & \begin{sideways}\textbf{Recall↑}\end{sideways} & \begin{sideways}\textbf{mAP$_{0.5}$↑}\end{sideways} & \begin{sideways}\textbf{mAP$_{0.5:0.95}$↑}\end{sideways} \\ 
\hline
Cows2021 & YOLOv8 \cite{YOLOv8} & 0.960 & 0.962 & 0.987 & 0.904 & N.A. & N.A. & N.A. & N.A. \\
Cows2021 & RT-DETR\cite{RT-DETR} & 0.981 & 0.950 & 0.989 & 0.845 & N.A. & N.A. & N.A. & N.A. \\
Ours & YOLOv8 \cite{YOLOv8} & 0.924 & 0.888 & 0.956 & 0.858 & 0.816 & 0.712 & 0.809 & 0.570 \\
Ours & RT-DETR\cite{RT-DETR} & 0.917 & 0.809 & 0.877 & 0.678 & 0.876 & 0.662 & 0.720 & 0.465 \\

\hline
\end{tabular}
}
\vspace{-0.2cm}
\end{table}

\Cref{tab:tracking} shows the results of tracking evaluated on a series of metrics, such as Precision, Recall, Multiple Object Tracking Accuracy (MOTA) and Precision (MOTP) that are used in MOT-Challenge \cite{MOTS} evaluation metrics, as well as Identification Precision (IDP), Recall (IDR), and F1 score (IDF1). The tracking models use the state-of-the-art YOLOv8 \cite{YOLOv8} as the detection model and are evaluated using py-motmetrics \cite{pymotmetrics}. The models achieve good MOTA and MOTP tracking scores, since the body detection models have already achieved good performance as seen from the high IDP, IDR, and IDF1 scores. The table also shows the accuracy of cow counts that is evaluated using Mean Absolute Error (MAE). 

\vspace{-0.2cm}
\begin{table}[ht]
\caption{Results of cattle tracking and counting}
\vspace{-0.2cm}
\resizebox{\linewidth}{!}{
\begin{tabular}{|c|c|c|c|c|c|c|c|c|} 
\hline
\textbf{Model} & 
\begin{sideways}\textbf{Precision↑}\end{sideways} & 
\begin{sideways}\textbf{Recall↑}\end{sideways} & 
\begin{sideways}\textbf{MOTA↑}\end{sideways} & 
\begin{sideways}\textbf{MOTP↑}\end{sideways} & 
\begin{sideways}\textbf{IDF1↑}\end{sideways} & 
\begin{sideways}\textbf{IDP↑}\end{sideways} & 
\begin{sideways}\textbf{IDR↑}\end{sideways} &
\begin{sideways}\textbf{MAE↓}\end{sideways} 
\\ 
\hline
BoTSORT \cite{BoT-SORT} & 0.211 & 0.036 & 0.167 & 0.312 & 0.025 & 0.087 & 0.015 & 20.5 \\
ByteTrack \cite{ByteTrack} & 0.201 & 0.034 & 0.170 & 0.336 & 0.025 & 0.091 & 0.015 & 24.0\\
\hline
\end{tabular}
\label{tab:tracking}
}
\vspace{-0.2cm}
\end{table}

While the cattle detection obtains promising results, it is still challenging to track the cow and obtain accurate count of the cows as seen in the low metric scores in \Cref{tab:tracking} and high MAE w.r.t. average of 58 cows in the videos (\ie, 41\% error) respectively. Tracking the cows is challenging as they may move back-and-forth across the tracking line, leading to multiple counts of the same cow. Also, the low scores for pose estimation in \Cref{tab:pose} is likely the result of the challenging top-down views. Hence, more advanced algorithms need to be developed to address these challenges.

\vspace{-0.2cm}
\begin{table}[ht]
\caption{Results of cattle pose estimation.}
\vspace{-0.2cm}
\label{tab:pose}
\centering
{
\begin{tabular}{|c|ccc|} 
\hline
\multirow{2}{*}{\textbf{Model}} & \multicolumn{3}{c|}{\textbf{OKS=0.5}}  \\ 
\cline{2-4}
 & \textbf{AP↑} & \textbf{AR↑} & \textbf{mAP↑}  \\ 
\hline
HRNet \cite{HRNet} & 0.459 & 0.615 & 0.122 
\\
YOLOv8 Pose \cite{YOLOv8} & 0.638 & 0.348 & 0.338  \\
\hline
\end{tabular}
}
\vspace{0.1cm}
\footnotesize{\\
OKS, AP, AR, mAP refer to Object Keypoint Similarity, Average Precision, Average Recall, mean Average Precision respectively.}
\vspace{-0.2cm}
\end{table}

The results in \Cref{tab:segmentation} shows that the models obtain high mAP$_{0.5}$ scores and are able to accurately segment the cows from the dirt background. 

\vspace{-0.2cm}
\begin{table}[h!]
\caption{Results of instance segmentation}
\vspace{-0.3cm}
\begin{center}
\begin{tabular}{|c|cccc|}
\hline
\textbf{Model} & \textbf{Precision↑} & \textbf{Recall↑} & \textbf{mAP$_{0.5}$↑} & \textbf{mAP$_{0.5:0.95}$↑} \\
\hline
YOLOv5 \cite{YOLOv5} &
0.877 & 0.774 & 0.875 & 0.661 \\
\hline
YOLOv8 \cite{YOLOv8} & 0.853 & 0.773 & 0.849 & 0.631 \\
\hline
\end{tabular}
\label{tab:segmentation}
\end{center}
\vspace{-0.2cm}
\end{table}

In summary, it can be observed that while good performance can be achieved for detection and segmentation, there remains a gap to achieve better performance. Moreover, it is still challenging to track the cow, estimate its keypoints, and obtain an accurate count. Thus, this shall motivate the community to develop, adapt, and evaluate various types of more advanced methods to tackle these challenges.

\section{Conclusion}
We construct CattleEyeView, the \textit{first} top-down view and multi-task cattle video dataset for a diversity of inter-related tasks (\ie, counting, detection, pose estimation, tracking, instance segmentation) that are useful to count the number of cows and assess the cow's growth and well-being. We perform benchmark experiments to evaluate the performance of models for each task. We hope that this dataset will facilitate the community and industry to develop, adapt, and evaluate various types of advanced methods to fulfill diverse needs and application scenarios for smarter precision livestock farming.

\section{Acknowledgments}
\footnotesize{This work was supported by the National Research Foundation Singapore under its AI Singapore Programme (Award number: Grant AISG-100E-2020-065). We would like to thank Coggan Farms for providing the video footages and making these footages available for research purposes, as well as our annotators at Keylabs.ai for annotating the videos and checking the quality of the annotations.
}

{\small
\bibliographystyle{ieee_fullname}
\bibliography{main.bib}
}
\end{document}